\documentclass[dvipsnames]{bmvc2k}
\usepackage{booktabs}
\usepackage{pifont}
\usepackage{amsfonts}
\usepackage{wrapfig}

\title{MOGRAS: Human Motion with Grasping in 3D Scenes}

\addauthor{Kunal Bhosikar}{https://kunal-kamalkishor-bhosikar.github.io/}{1}
\addauthor{Siddharth Katageri}{https://siddharthkatageri.github.io/}{1}
\addauthor{Vivek Madhavaram}{https://vivekmadhavaram.github.io/vivek_page/}{1}
\addauthor{Kai Han}{https://www.kaihan.org/}{2}
\addauthor{Charu Sharma}{https://charusharma.org/}{1}

\addinstitution{
 Machine Learning Lab,\\
 International Institute of Information Technology,\\
 Hyderabad, India
}
\addinstitution{
 Vision AI Lab,\\
 The University of Hong Kong,\\
 Pokfulam Road, Hong Kong
}

\runninghead{Bhosikar \etal}{MOGRAS}

\def\etal{\emph{et al}\bmvaOneDot}

\begin{document}

\maketitle

\begin{figure*}[h!]
    \centering
    \includegraphics[width=0.9\linewidth, alt={Teaser}, trim=1cm 1cm 1cm 1cm]{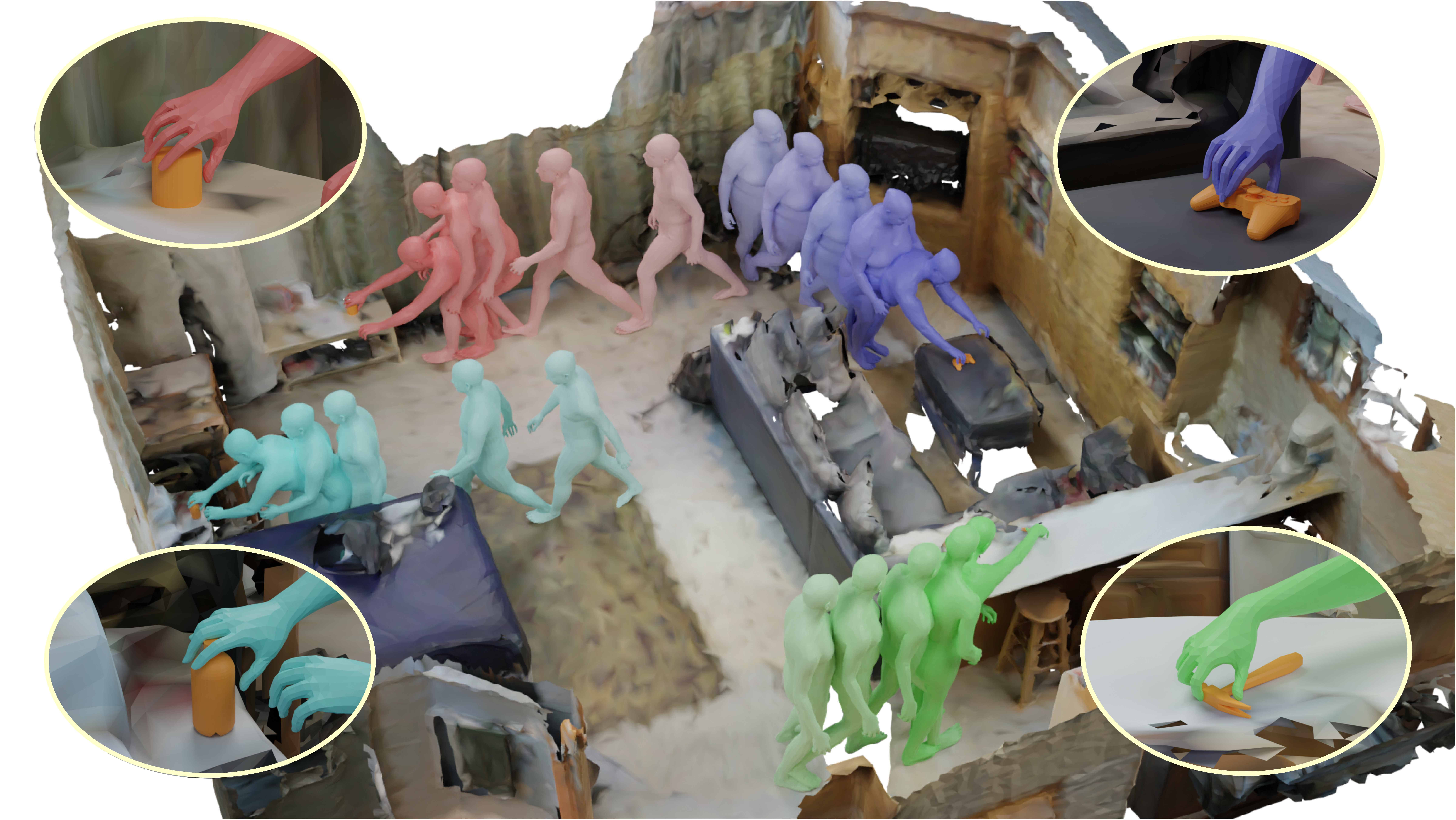}
    \caption{\textbf{Overview of the MOGRAS Dataset.} Our MOGRAS dataset provides full-body motion sequences, including both pre-grasp walking and detailed grasping poses, within richly annotated 3D scenes. This unique combination addresses the limitations of existing datasets, bridging the gap between synthetic data and real-world scenarios to enable realistic and complex human-object interactions.}
    \label{fig:teaser}
    \vspace{-1pt}
\end{figure*}

\begin{abstract}
Generating realistic full-body motion interacting with objects is critical for applications in robotics, virtual reality, and human-computer interaction. While existing methods can generate full-body motion within 3D scenes, they often lack the fidelity for fine-grained tasks like object grasping. Conversely, methods that generate precise grasping motions typically ignore the surrounding 3D scene. This gap, generating full-body grasping motions that are physically plausible within a 3D scene, remains a significant challenge. To address this, we introduce \textbf{MOGRAS} (Human \textbf{MO}tion with \textbf{GRA}sping in 3D \textbf{S}cenes), a large-scale dataset that bridges this gap. MOGRAS provides pre-grasping full-body walking motions and final grasping poses within richly annotated 3D indoor scenes. We leverage MOGRAS to benchmark existing full-body grasping methods and demonstrate their limitations in scene-aware generation. Furthermore, we propose a simple yet effective method to adapt existing approaches to work seamlessly within 3D scenes. Through extensive quantitative and qualitative experiments, we validate the effectiveness of our dataset and highlight the significant improvements our proposed method achieves, paving the way for more realistic human-scene interactions.
\end{abstract}

\section{Introduction}
\label{sec:intro}
Understanding how humans interact with objects in 3D space is fundamental to numerous domains, including computer vision, robotics, animation, and virtual reality~\cite{detry2010refining, Hsiao2006ImitationLO, li2007pruning, seo2012wholearm, kry2006synthesis}. Accurately modeling these interactions enables behavior analysis, the development of intelligent robotic systems, and the creation of realistic, immersive virtual environments~\cite{rijpkema1991grasping, kalisiak2001animation}. A key challenge lies in simulating realistic and physically plausible human-object interactions (HOI), which involves modeling contact, force, and object manipulation in complex 3D scenes~\cite{Krug445289}.

Despite recent advances, capturing the nuanced dynamics of HOI remains difficult. While many existing datasets and methods focus on large-scale interactions like sitting on a sofa or using a bed~\cite{wang2022humanise}, others target fine-grained hand-object manipulation, often ignoring the rest of the body~\cite{korrawe2020graspingfield, javier2022embodiedhands, hanwen2021generation, patrick2021contactopt}. Some work has extended this to full-body grasping~\cite{GRAB:2020}, but these approaches typically neglect scene context. This highlights a critical gap: the lack of datasets and models that account for full-body grasping of small objects within real 3D indoor scenes. Addressing this is essential for advancing scene-aware HOI, as performing full-body motion with objects in cluttered environments requires precision in contact and collision handling. 

To address this challenge, we introduce MOGRAS, a novel, large-scale synthetic dataset containing full-body motion sequences and grasping interactions in indoor 3D scenes. Each sequence includes both pre-grasp walking motion and the final grasping pose, as shown in Figure~\ref{fig:teaser}. Our pipeline ensures minimal collision between the human mesh, the grasped object, and the surrounding scene geometry, enabling realistic motion and contact.

Manually capturing such complex HOI data is prohibitively expensive, requiring specialized hardware and extensive manual labor. To overcome this, we propose an automated data generation framework comprising five key steps: $(i)$ placing small objects on surfaces, $(ii)$ aligning pre-grasp walking motion, $(iii)$ refining the scene to handle collisions, $(iv)$ generating a realistic grasping pose, and $(v)$ infilling the motion to smoothly transition to the grasp. Given the synthetic nature of our dataset, we conduct an extensive human evaluation to validate its realism and naturalness, ensuring its suitability for training models for multiple downstream tasks.

We benchmark existing grasping models, GOAL~\cite{taheri2021goal} and SAGA~\cite{wu2022saga}, and observe that they fail to produce plausible grasps without significant scene penetration. To address this, we extend the GOAL's GNet~\cite{taheri2021goal} architecture by introducing a novel penetration loss and conditioning the model on the 3D scene mesh. This modification, which we name GNet++, significantly improves grasping quality in cluttered scenes. Our work emphasizes the importance of scene-aware grasping, where the objective is to minimize the intersection between the human body and the scene. We show how our method uses a pretrained model to generate initial grasps and refines them using our proposed loss to reduce collisions and produce physically plausible contact.

Our contributions can be summarized as follows: \\
$(i)$ We introduce MOGRAS, a large-scale synthetic dataset for scene-aware, full-body grasping. It includes both pre-grasp approaching motion and final grasping poses, generated automatically through our proposed framework. \\
$(ii)$ We provide a comprehensive benchmark of existing grasping methods on our challenging dataset, highlighting their limitations. \\
$(iii)$ We propose a novel method, GNet++, that improves scene-aware grasping by leveraging our dataset and introducing a new penetration loss to produce physically plausible results.

\section{Related Work}
\label{sec:related}

Our work is positioned at the intersection of 3D human-object grasping and human-scene interaction. We review key advancements in both areas to highlight the specific gap our work addresses.

\textbf{3D Human-Object Grasping.}
Research in 3D grasp synthesis has historically focused on isolated hand-object interactions. Generative models like Grasping Field~\cite{korrawe2020graspingfield} and GrabNet~\cite{GRAB:2020} used implicit functions and VAEs to produce realistic hand poses. Subsequent work, such as GraspTTA~\cite{hanwen2021generation} and ContactOpt~\cite{patrick2021contactopt}, refined these methods through contact-aware optimization to improve physical plausibility.

More recently, the focus has expanded to full-body grasp synthesis, leveraging datasets like GRAB~\cite{GRAB:2020}, which provides MoCap data of SMPL-X~\cite{pavlakos2019smplx} bodies interacting with objects. Models like GOAL~\cite{taheri2021goal} and SAGA~\cite{wu2022saga} have built on this to generate full-body poses for handheld objects. Even more recent works, such as DiffGrasp~\cite{zhang2024diffgraspwholebodygraspingsynthesis}, explore diffusion-based approaches for generating high-quality full-body grasping sequences. However, a critical limitation of these methods is their lack of scene awareness; they generate grasps in isolation, ignoring potential collisions with the surrounding environment. While OmniGrasp~\cite{luo2025omnigraspgraspingdiverseobjects} explores diverse object grasping with virtual humanoids, it also omits real-world scene context.

A few methods have attempted to incorporate scene constraints. FLEX~\cite{tendulkar2022flex} generates scene-aware full-body grasps by leveraging hand-object and foot-ground contacts, but its optimization-based inference is computationally expensive. VirtualHome~\cite{Puig2018VirtualHomeSH} simulates household activities but often produces unrealistic interactions with significant penetration due to its reliance on programmatic, pre-defined animations. These approaches show the need for scene context but highlight the difficulty of achieving it efficiently and realistically.

\textbf{3D Human-Scene Interaction (HSI).}
Parallel research in human-scene interaction (HSI) aims to generate plausible human motions within rich 3D environments. Early optimization-based methods used physical priors and contact annotations to place humans in scenes~\cite{kim2014shape2pose, zheng2016ergonomics, kurt2020posetoseat}. Modern learning-based approaches have improved scalability by conditioning pose generation on scene representations, using either scene-centric~\cite{zhang2020place} or human-centric~\cite{Hassan_2021_CVPR} perspectives.

Large-scale datasets have been pivotal in advancing HSI. Datasets like HUMANISE~\cite{wang2022humanise}, GTA-IM~\cite{cao2020longterm}, and more recent works like SMPLOlympics~\cite{luo2024smplolympicssportsenvironmentsphysically} and UniHSI~\cite{xiao2024unifiedhumansceneinteractionprompted} provide diverse human motions in varied environments. TRUMANS~\cite{jiang2024scaling} further improves data quality with high-fidelity contact supervision. However, the focus of these HSI datasets is on large-scale interactions like sitting, walking, or navigating. They generally lack the fine-grained contact information and specific intent required for precise object grasping. Consequently, models trained on them excel at general scene placement but fail at detailed manipulation tasks.

\textbf{Positioning and Contribution.}
The existing literature reveals a clear dichotomy: 3D grasping research is largely scene-agnostic, while 3D HSI research focuses on large-scale interactions and lacks fine-grained grasping detail. This leaves a critical, unaddressed gap for tasks requiring full-body, scene-aware manipulation of small objects. Our work directly targets this gap. MOGRAS is the first large-scale dataset to provide physically plausible, full-body grasping motions within cluttered indoor scenes. By synthesizing realistic approaching motions and collision-free final grasps, we provide the necessary data to train and evaluate the next generation of truly context-aware HOI models.

\section{The MOGRAS Dataset}
\label{dataset}

Existing datasets for full-body human-object interaction (HOI)~\cite{bhatnagar2022behave,wang2023physhoi,li2023object,tendulkar2022flex} suffer from several limitations. As we outlined in the previous section, datasets capturing human motion in 3D environments tend to focus on interactions with large, static objects like tables or sofas, while overlooking fine-grained interactions with small, handheld objects. Conversely, datasets for small object interaction often omit the 3D scene context entirely, failing to account for critical environmental constraints. Since scene context fundamentally shapes how humans approach, manipulate, and grasp objects, a realistic model of full-body HOI must incorporate both fine-grained interactions and the surrounding 3D scene.

\begin{figure}[ht]
  \centering
  \includegraphics[width=0.9\linewidth, alt={Dataset}, trim=6cm 3cm 6cm 0cm]{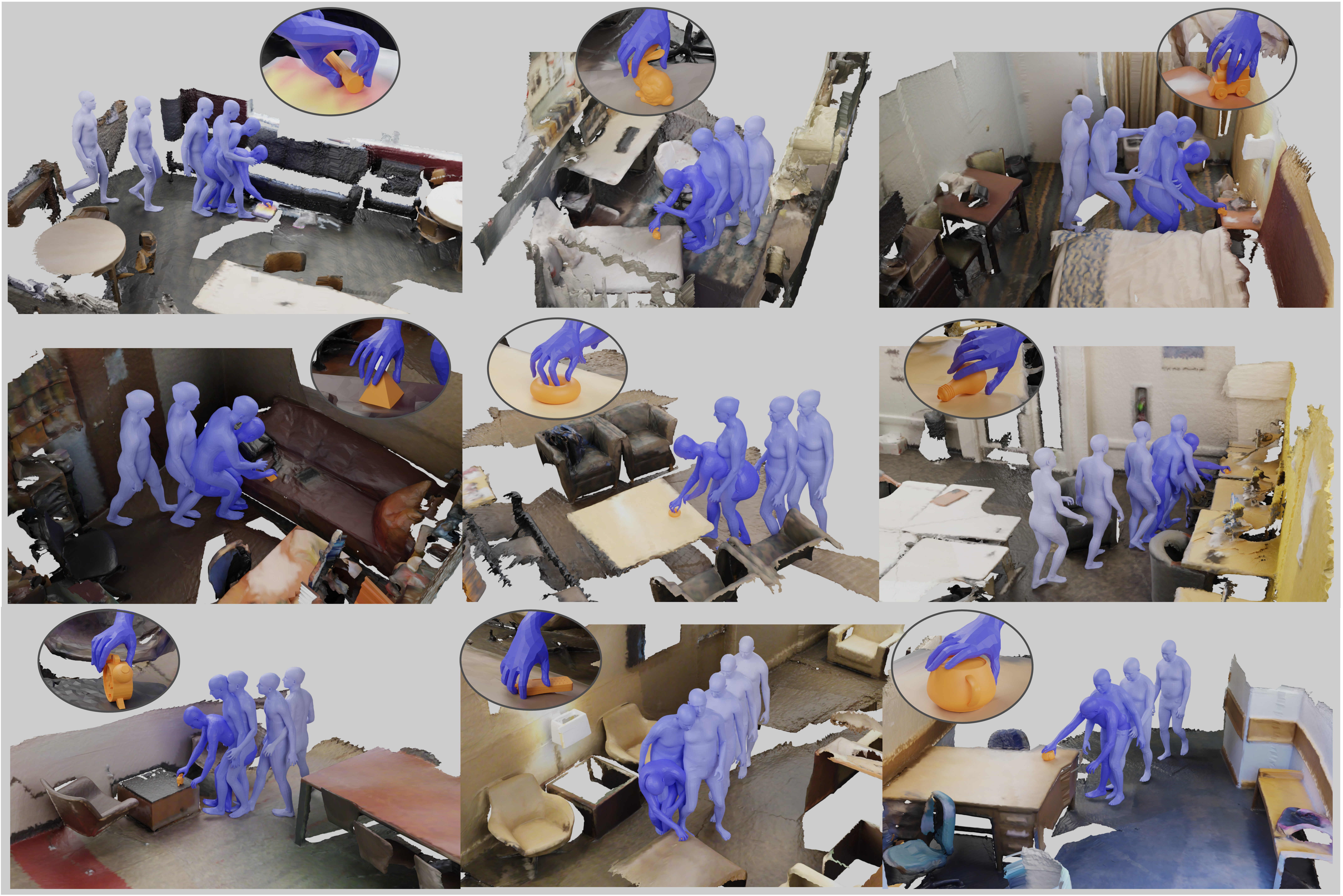}
  \caption{\textbf{Examples from the MOGRAS dataset.} MOGRAS provides full-body motion sequences, including a pre-grasp walking phase and a final grasping pose, all within richly annotated 3D scenes. The dataset captures subtle, physically plausible interactions with small objects and their environment, addressing a key gap in existing research.}
  \label{fig:dataset}
  \vspace{-19pt}
\end{figure}

To address these shortcomings, we introduce MOGRAS, a large-scale, synthetic dataset featuring full-body grasping in diverse 3D indoor scenes. Manually acquiring such a dataset would be prohibitively expensive and time-consuming. To overcome this, we propose an automated synthesis pipeline that efficiently generates high-quality interaction sequences. The core idea is to first align an existing human walk motion to a 3D scene, then synthesize a physically plausible grasping pose, and finally infill the motion to smoothly transition from the walk to the grasp. Figure~\ref{fig:dataset} provides an overview of our data generation process. 

Our pipeline consists of five key steps, which we detail below.

\subsection{Walk Motion Alignment and Object Placement}
\label{subsec:alignment_and_placement}
We follow the methodology of HUMANISE~\cite{wang2022humanise} to align walking motions with 3D scenes. We use walking sequences from AMASS~\cite{mahmood2019amass} and 3D environments from ScanNetv2~\cite{dai2017scannet}. Motion labels from BABEL~\cite{BABEL:CVPR:2021} are used to segment walk motions, while graspable objects are sourced from GRAB~\cite{GRAB:2020}. Our dataset includes 2,988 walking motion clips, each 1 to 4 seconds in duration. 

For each sequence, we first identify a scene element, or \emph{receptacle} (e.g., a table, counter, or desk), from ScanNet's~\cite{dai2017scannet} semantic annotations. We then optimize for a global translation \(t\) and rotation \(R\) to ensure the human body is collision-free and the walking trajectory terminates near the chosen receptacle~\cite{wang2022humanise}. 

Once the motion is aligned, we place a graspable object on the receptacle's surface to ensure it is reachable from the final human position. We achieve this by extracting the receptacle mesh, identifying a suitable placement point using a KD-tree, and placing the object at the highest point on the surface to prevent floating artifacts. This pipeline generalizes well across diverse scenes with high-quality semantic segmentation (see Figure ~\ref{fig:dataset}).

\subsection{Refining Scenes from ScanNet}
\label{subsec:refining_scene}

\begin{wrapfigure}{r}{0.5\linewidth}
  \centering
  \includegraphics[width=0.9\linewidth, alt={Floor}, trim=5cm 7cm 5cm 7cm]{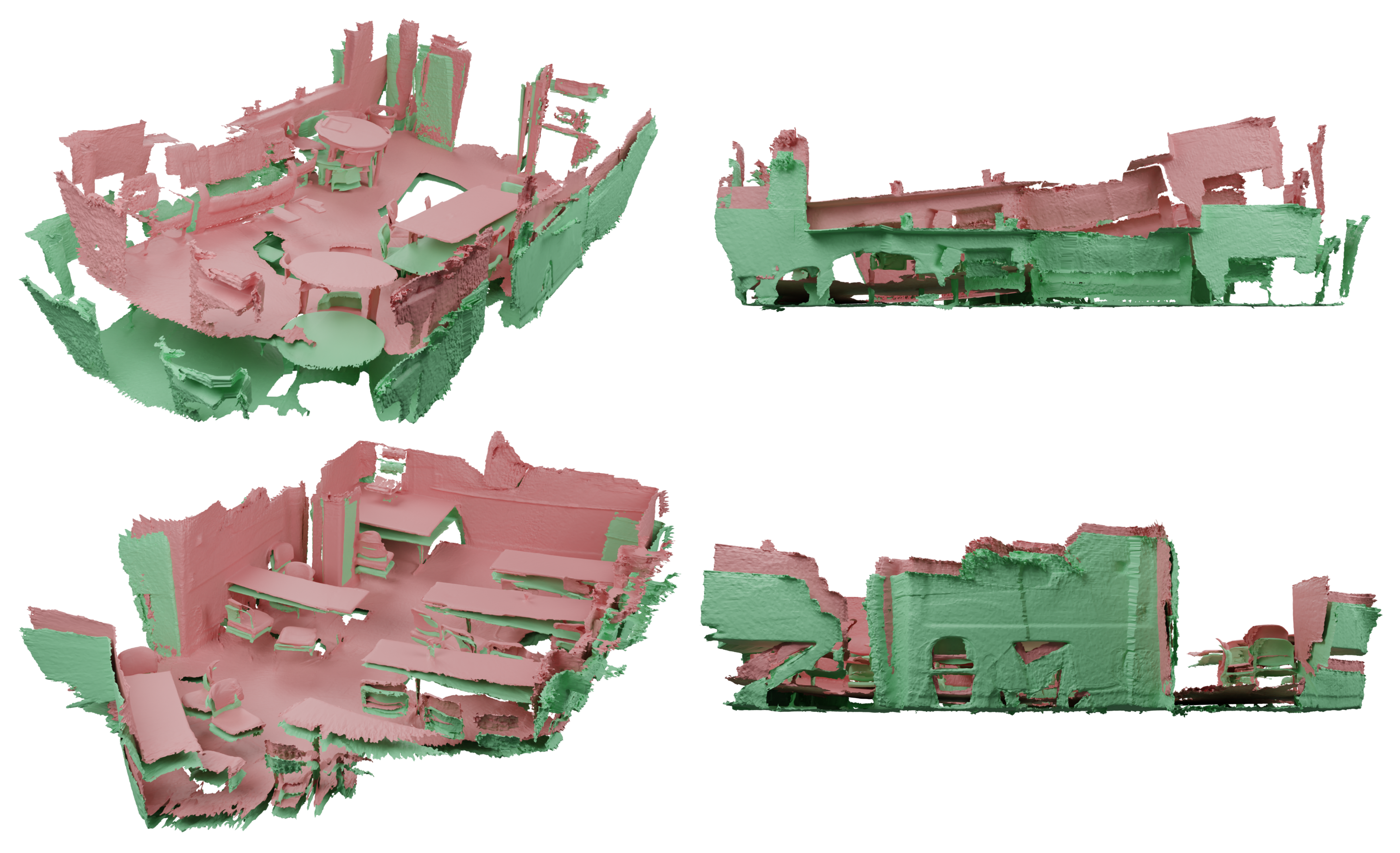}
  \caption{\textbf{Comparison of Original and Refined ScanNet Scenes.} Original scenes (pink) often have floor misalignment, causing artifacts like floating feet. Our refined scenes (green) correct this, improving physical plausibility and interaction quality.}
  \label{fig:scene_refine}
  \vspace{-13pt}
\end{wrapfigure}

ScanNet~\cite{dai2017scannet} scenes often suffer from floor misalignment with the ideal \(z=0\) ground plane, leading to foot-floating or penetration artifacts in the generated motion. A global rigid transform is inadequate due to spatially varying deviations, while non-rigid warping risks distorting object layouts. To address this, we propose a piecewise rigid alignment method that adjusts floor geometry while preserving the overall scene structure (see Figure~\ref{fig:scene_refine}). 

\begin{wraptable}{r}{\linewidth}
    \centering
    \caption{\textbf{Quantitative Floor Alignment.} This table shows that our method significantly reduces the deviation of floor vertices from the ground plane ($z=0$) in refined ScanNet scenes, demonstrating improved alignment.}
    \label{table:scene_refinement_stats}
    \resizebox{0.5\columnwidth}{!}{%
    \begin{tabular}{l|c|c|c|c}
        \toprule
        & \textbf{Average} & \textbf{Average} & \textbf{Global Mean} & \textbf{Global} \\
        & \textbf{Mean} & \textbf{Standard} & \textbf{Mean} & \textbf{Standard} \\
        & & \textbf{Deviation} & & \textbf{Deviation} \\
        \midrule
        \textbf{Original} & 0.1175 & 0.1303 & 0.1396 & 0.2591 \\
        \textbf{Refined} & \textbf{0.0045} & \textbf{0.0208} & \textbf{0.0039} & \textbf{0.0394} \\
        \bottomrule
    \end{tabular}%
    }
\end{wraptable}

Our solution applies local rigid transforms computed via ICP over fixed-size sliding windows on the \(x-y\) plane. For each window, we extract floor vertices and align them to the \(z=0\) plane by optimizing for vertical translation \(\mathsf{t}_z\) and in-plane rotations \(\mathsf{R}_x, \mathsf{R}_y\). This strategy yields significantly better floor alignment, as quantified in Table~\ref{table:scene_refinement_stats}, and enables more physically plausible human-scene interactions.

\subsection{Generating Grasping Poses}
\label{subsec:gen_grasp_pose}

We generate realistic full-body grasping poses conditioned on object placement and scene context using FLEX~\cite{tendulkar2022flex}, an optimization-based method that searches the latent spaces of VPoser~\cite{vposer} (body pose) and GrabNet~\cite{GRAB:2020} (hand pose). Although FLEX~\cite{tendulkar2022flex} yields high-quality poses, its computational cost is high. To enable large-scale generation, we adapt FLEX~\cite{tendulkar2022flex} by reducing optimization iterations and applying two post-optimization refinements, achieving a significant speedup without compromising pose quality. We provide implementation details of these modifications in the Supplementary Appendix.

\subsection{Generating Final Approaching Motion}
\label{subsec:infill_approach_motion}

To create a coherent sequence, we seamlessly integrate the generated grasping pose (Section~\ref{subsec:gen_grasp_pose}) with the preceding walking motion (Section~\ref{subsec:alignment_and_placement}) using PriorMDM~\cite{priormdm}, a motion infilling model. We adopt its root-trajectory variant, which generates intermediate poses by interpolating between the pelvis of the final walking frame and the grasping pose. To avoid unnatural backtracking, we adjust the grasping location to ensure a forward motion, preserving trajectory realism. The infill duration scales with the distance between the start and end poses, allowing for smooth, adaptive transitions. Additionally, for each object-receptacle pair, we provide both the generated motion and multiple pelvis locations near the grasp. These serve as valid, collision-free grasp targets to support downstream tasks like evaluation or learning.

\subsection{Dataset Scale and Quality}
\label{subsec:data_stats}

\begin{wraptable}{r}{0.5\linewidth}
    \centering
    \caption{\textbf{Statistics of the MOGRAS Dataset Splits.} This table presents a breakdown of the training, validation, and test sets. The splits are structured to ensure diversity and broad coverage of scenes and object interactions.}
    \label{table:stats}
    \resizebox{0.47\columnwidth}{!}{%
    \begin{tabular}{l|c|c|c|c}
        \toprule
        \textbf{Split} & \textbf{\#Motion} & \textbf{\#Frames} & \textbf{\#Scenes} & \textbf{\#Object} \\
        & \textbf{Samples} & & & \textbf{Instances} \\
        \midrule
        \textbf{Train} & 11,479 & 678,226 & 427 & 37 \\
        \textbf{Val}   &    932 &  54,484 &  27 &  4 \\
        \textbf{Test}  &  1,827 & 106,079 &  53 &  6 \\
        \bottomrule
    \end{tabular}%
    }
\end{wraptable}

MOGRAS is a large-scale, high-quality dataset for full-body human-object interaction. As detailed in Table \ref{table:stats}, MOGRAS contains over 14k motion sequences across 500 scenes and 47 objects. Unlike prior work that focuses on either large-scale interactions or scene-agnostic grasping, MOGRAS uniquely combines full-body motion, fine-grained grasping, and 3D scene context (Table \ref{table:dataset_comparison}).

\begin{table}[ht]
    \centering
    \caption{\textbf{Comparison of Dataset Statistics and Functional Coverage.} This table summarizes existing datasets against MOGRAS across key dimensions: the number of samples, frames, scenes, and objects. We also compare support for three critical capabilities: 3D scene inclusion, fine-grained grasping, and full-body motion. MOGRAS is the only dataset to combine all three, addressing a critical gap in the literature.}
    \label{table:dataset_comparison}
    \resizebox{\columnwidth}{!}{%
    \begin{tabular}{l|c|c|c|c|c|c|c}
        \toprule
        \textbf{Dataset} & \textbf{\#Samples} & \textbf{\#Frames} & \textbf{\#Scenes} & \textbf{\#Objects} & \textbf{3D}  & \textbf{Fine-} & \textbf{Full-} \\
         & & & & & \textbf{Scene} & \textbf{Grained} & \textbf{Body} \\
         & & & & & \textbf{Inclusion} & \textbf{Grasping} & \textbf{Motion} \\
        \midrule
        \textbf{HUMANISE}~\cite{wang2022humanise}  & 19.6k & 1.2M & 643 & {\color{BrickRed}\ding{55}} & {\color{ForestGreen}\ding{51}} & {\color{BrickRed}\ding{55}} & {\color{ForestGreen}\ding{51}}\\
        \textbf{GRAB}~\cite{GRAB:2020} & 1.3k & 1.6M & {\color{BrickRed}\ding{55}} & 51 & {\color{BrickRed}\ding{55}} & {\color{ForestGreen}\ding{51}} & {\color{ForestGreen}\ding{51}}\\
        \textbf{OakInk}~\cite{yang2022oakinklargescaleknowledgerepository} & 50k & 0.23M & {\color{BrickRed}\ding{55}} & 1800 & {\color{BrickRed}\ding{55}} & {\color{ForestGreen}\ding{51}} & {\color{BrickRed}\ding{55}}\\
        \textbf{SceneFun3D}~\cite{delitzas2024scenefun3d} & 14.8k & {\color{BrickRed}\ding{55}} & 710 & {\color{BrickRed}\ding{55}} & {\color{ForestGreen}\ding{51}} & {\color{BrickRed}\ding{55}} & {\color{BrickRed}\ding{55}}\\
        \textbf{ScenePlan}~\cite{xiao2024unifiedhumansceneinteractionprompted} & 1.1k & {\color{BrickRed}\ding{55}} & 10 & 40 & {\color{ForestGreen}\ding{51}} & {\color{BrickRed}\ding{55}} & {\color{ForestGreen}\ding{51}}\\
        \textbf{MOGRAS} & 14.2k & 0.8M & 507 & 47 & {\color{ForestGreen}\ding{51}} & {\color{ForestGreen}\ding{51}} & {\color{ForestGreen}\ding{51}}\\
        \bottomrule
    \end{tabular}%
    }
\end{table}

To ensure realism, we use a multi-stage quality assurance process that includes automated penetration filtering and a human study. In this study, 75 participants rated our motions for naturalness and interaction realism, yielding high scores of 4.2/5 and 4.5/5, respectively. Motions below a rating of 3.5 were removed, ensuring our synthetic data is physically plausible and suitable for training models.

\section{Scene-Aware Grasping with GNet++}
\label{sec:benchmarking}
We benchmark our MOGRAS dataset on the task of scene-aware grasping, focusing specifically on the static grasp pose rather than full motion sequences. This decision reflects the fine-grained challenge of generating physically plausible grasps in complex 3D scenes, where avoiding human-scene and human-object intersections is critical. Our proposed method, GNet++, is an extension of the pretrained GNet~\cite{taheri2021goal} model, which we use as our baseline to refine scene-agnostic grasps into context-aware and physically valid ones.

\subsection{Preliminaries: GOAL and GNet}
\label{subsec:prelim}
\textbf{SMPL-X}~\cite{pavlakos2019smplx} is a parametric 3D human model with 10,475 vertices, controlled by shape, pose, and expression parameters. It is widely used to represent human bodies, faces, and hands in a unified mesh.

\textbf{GOAL}~\cite{taheri2021goal} is a two-stage framework for human-object interaction. Given an object and an initial human pose, it first uses GNet to predict a realistic whole-body grasping pose. The second stage, MNet, then generates a full motion trajectory. Our work focuses on improving the GNet stage.

\textbf{GNet} is a conditional variational autoencoder (cVAE) that generates full-body grasp poses conditioned on object geometry and position. Its encoder maps input parameters (SMPL-X~\cite{pavlakos2019smplx}, BPS~\cite{sergey2019bps} object encoding, and object translation) to a latent space. The decoder uses a sampled latent grasp code and object condition to predict SMPL-X~\cite{pavlakos2019smplx} parameters, a head orientation, and offset vectors for refining hand-object contact.

\subsection{Our Method: GNet++}
\label{subsec:method}

We introduce GNet++, a scene-aware extension of GNet, as illustrated in Figure~\ref{fig:arch}. We make two key modifications to enable it to reason about environmental constraints:

\begin{wrapfigure}{r}{0.5\linewidth}
\centering
\includegraphics[width=\linewidth, alt={Architecture}]{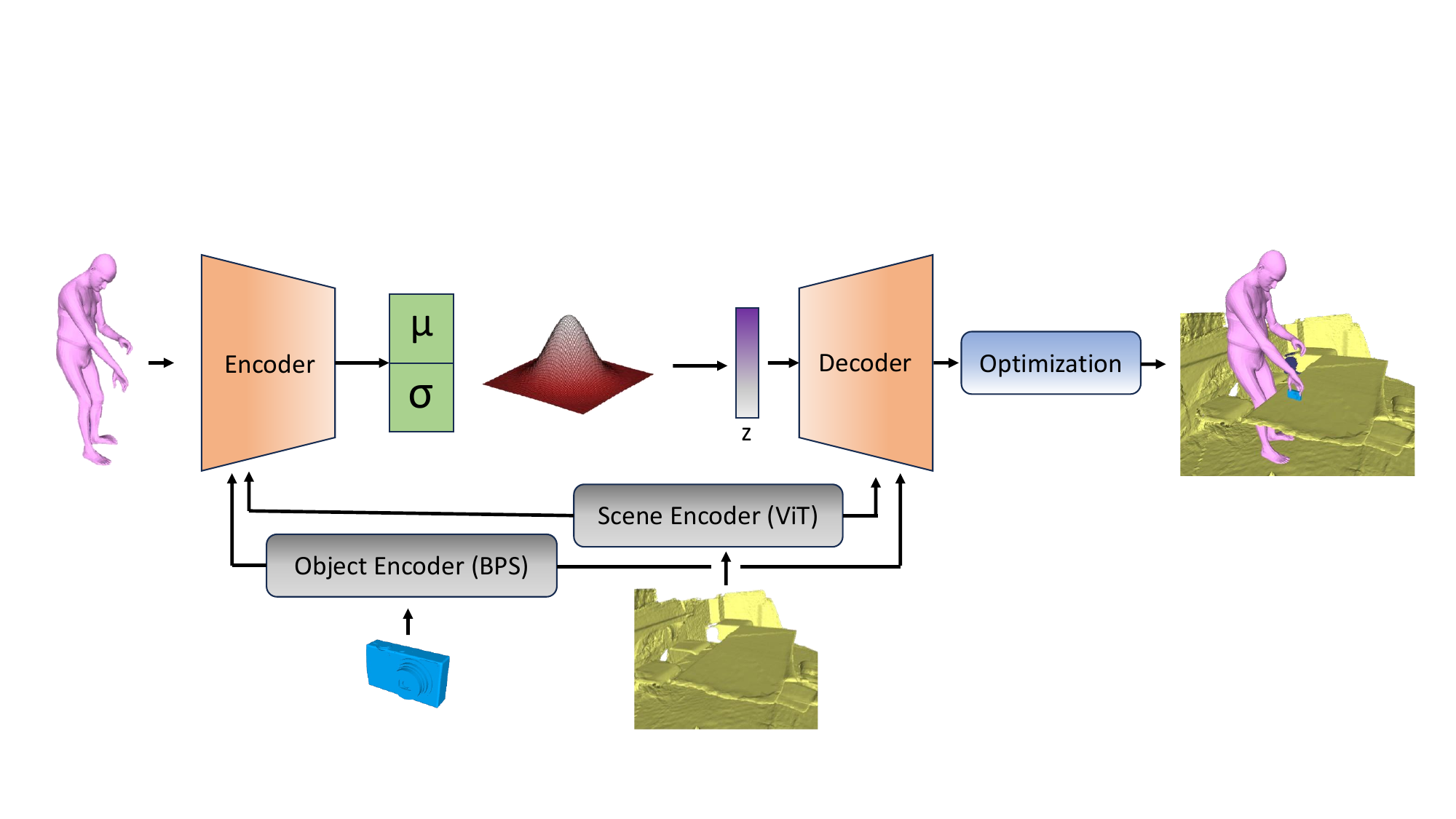}
\caption{\textbf{GNet++ Architecture Overview.} We extend the GNet architecture to create GNet++ by injecting a scene embedding into both the encoder and decoder, enabling it to generate context-aware grasps that avoid collisions.}
\label{fig:arch}
\end{wrapfigure}

\textbf{1. Scene-Conditioned Architecture:} We augment GNet's input to include scene information. The 3D scene is encoded into a feature vector using a pretrained Vision Transformer (ViT)~\cite{vit} on a bird's-eye-view (BEV) projection. This scene embedding is fused with the object encoding and injected into both the encoder and decoder of the cVAE, allowing the model to learn the relationship between the grasp pose and the surrounding geometry. While the BEV projection effectively captures floor-level and planar obstacles, we acknowledge its limitation in fully representing vertical or overhanging obstacles (e.g., shelves) that are not directly projected onto the ground plane. Our proposed penetration loss helps mitigate this by explicitly penalizing human-scene intersections in 3D space.

\textbf{2. Penetration-Aware Loss:} To explicitly discourage collisions, we introduce a new loss term, \((\mathcal{L}_{pen})\). We first voxelize the 3D scene around the object. Any voxel below an occupied one is also marked as occupied to account for unreachable spaces. During training, we penalize any predicted human mesh vertex that falls within these occupied voxels. This loss, combined with fine-tuning on our MOGRAS dataset, teaches the model to generate physically plausible poses that respect scene boundaries.

\subsection{Experiments and Evaluation}
We evaluate GNet++ on the test set of MOGRAS, comparing it against the original GOAL~\cite{taheri2021goal}, SAGA~\cite{wu2022saga}, and ablations of our own model. For a comprehensive overview of our baseline evaluation and experimental setup, please refer to the Appendix.

\noindent\textbf{Metrics.} To assess physical plausibility and interaction realism, we report performance on four key metrics:

\textbf{Human-Scene Penetration\((\downarrow)\):} The proportion of human body vertices located inside occupied voxels of the 3D scene, indicating undesired interpenetrations.

\textbf{Human-Floor Penetration\((\downarrow)\):} The fraction of foot vertices located below the floor plane, penalizing floating or sinking artifacts.

\textbf{Human-Object Penetration\((\downarrow)\):} Mean signed distance function (SDF) values computed at hand vertices.

\textbf{Contact Precision/Recall\((\uparrow)\):} F1 score measuring how accurately predicted contact points align with ground-truth contact areas on objects and the floor. For a detailed definition of our ground-truth contact annotations, please see the Appendix.

\begin{figure}[ht]
    \centering
    \includegraphics[width=\linewidth]{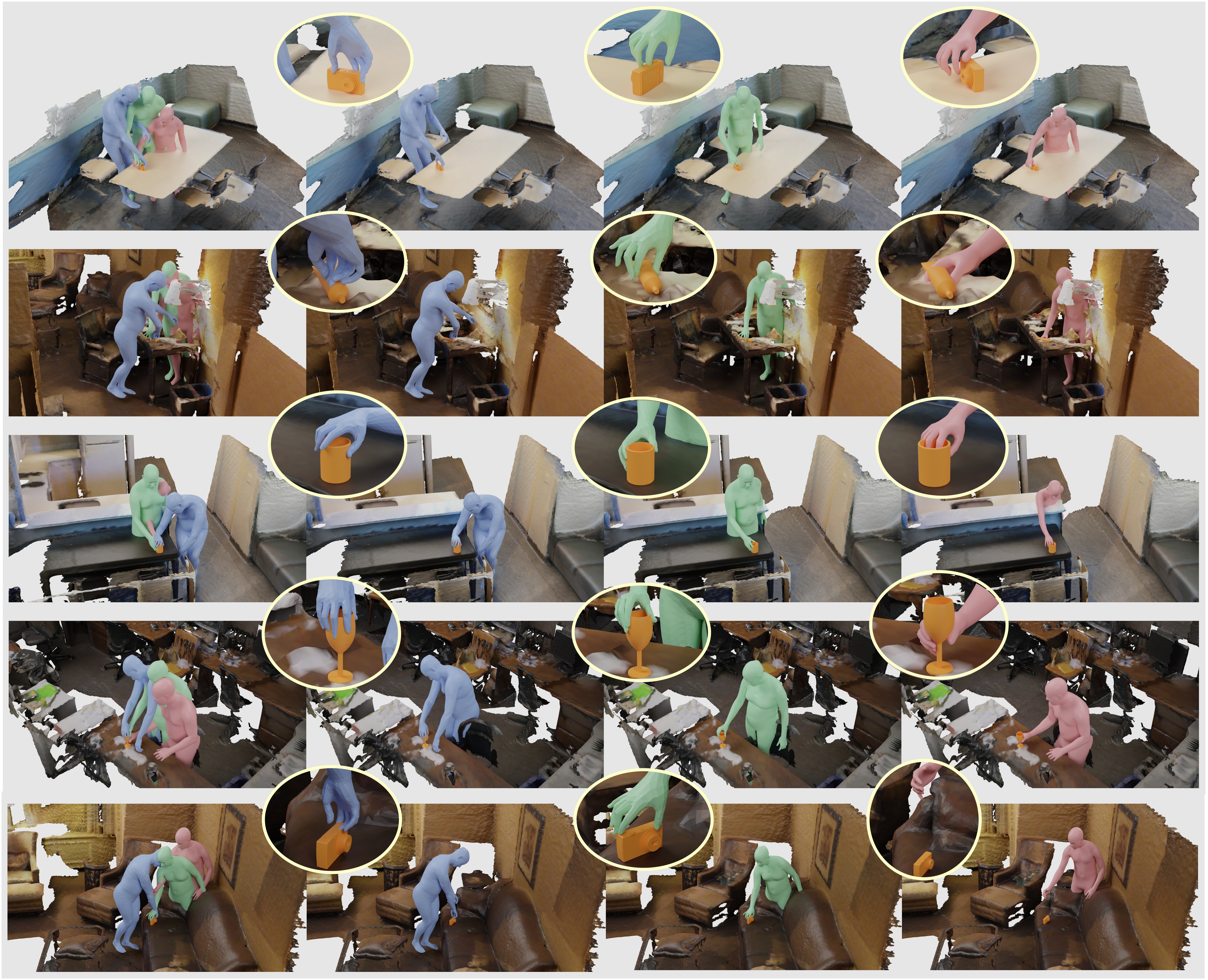}
    \caption{\textbf{Qualitative Comparison of Generated Grasps.} We visualize grasp poses from GNet++ (blue), GOAL (green), and SAGA (pink). In these cluttered scenes, both GOAL and SAGA produce significant penetration with the table and surrounding objects. In contrast, GNet++ successfully generates a physically plausible and collision-free pose by adapting the body to the environment.}
    \label{fig:result_compare}
\end{figure}

\noindent\textbf{Quantitative Results.} Table~\ref{table:eval} shows that GNet++ significantly outperforms all baselines. Compared to GOAL, GNet++ reduces scene penetration by 28\% and object penetration by 58\%, while dramatically improving floor contact (0.92 vs. 0.76 F1). Figure~\ref{fig:result_compare} visually confirms these findings, showing that GOAL and SAGA frequently generate poses that intersect with the table, whereas GNet++ produces a realistic, collision-free crouch.

\begin{table}[ht]
    \centering
    \caption{\textbf{Quantitative Comparison on Scene-Aware Grasping.} We evaluate penetration \((\downarrow)\) and contact quality \((\uparrow)\). Our full GNet++ model, fine-tuned on MOGRAS with the proposed penetration loss, significantly outperforms both state-of-the-art baselines and its ablated variants across all metrics. This demonstrates the importance of both scene-aware conditioning and explicit collision penalization.}
    \label{table:eval}
    \resizebox{\columnwidth}{!}{%
    \begin{tabular}{l|ccc|ccc|ccc}
        \toprule
        & \multicolumn{3}{c|}{\textbf{Penetration} $(\downarrow)$} & \multicolumn{3}{c|}{\textbf{Object Contact} $(\uparrow)$} & \multicolumn{3}{c}{\textbf{Floor Contact} $(\uparrow)$} \\
        \textbf{Method} & Scene & Object & Floor & Precision & Recall & F1 & Precision & Recall & F1 \\
        \midrule
        GOAL~\cite{taheri2021goal} & 4.35\% & 3.49\% & 3.62\% & 0.85 & 0.76 & 0.80 & 0.70 & 0.83 & 0.76 \\
        SAGA~\cite{wu2022saga} & 5.80\% & 2.67\% & 1.14\% & 0.77 & 0.71 & 0.74 & 0.81 & 0.83 & 0.82\\
        \midrule
        GNet++ (w/o MOGRAS ft) & 4.47\% & 2.48\% & 0.58\% & 0.85 & 0.77 & 0.81 & 0.72 & 0.84 & 0.78\\
        GNet++ (w/o pen. loss) & 3.91\% & 2.07\% & 0.098\% & 0.74 & 0.70 & 0.72 & 0.81 & 0.88 & 0.84 \\
        \textbf{GNet++ (Ours)} & \textbf{3.13\%} & \textbf{1.45\%} & \textbf{0.029\%} & \textbf{0.86} & \textbf{0.79} & \textbf{0.82} & \textbf{0.90} & \textbf{0.95} & \textbf{0.92}\\
        \bottomrule
    \end{tabular}%
    }
\end{table}

\noindent\textbf{Ablation Studies.} Our ablations in Table~\ref{table:eval} isolate the impact of our key contributions.

\textbf{Effect of MOGRAS:} Fine-tuning on our dataset (comparing row 3 vs. row 5) is crucial for reducing floor and scene penetration, as it provides the necessary examples of full-body interaction in context.

\textbf{Effect of Penetration Loss:} Adding our explicit penetration loss (comparing row 4 vs. row 5) provides the largest single improvement in scene penetration, proving its effectiveness in enforcing physical constraints.

These results confirm that both our MOGRAS dataset and the proposed GNet++ architecture are essential for achieving high-quality, scene-aware grasping.

\section{Conclusion}
In this work, we introduced MOGRAS, a large-scale dataset that addresses a critical gap in human-scene interaction research by providing full-body grasping motions within complex 3D scenes. We also proposed GNet++, a novel generative model that leverages this dataset to produce physically plausible and scene-aware grasps. Our extensive quantitative and qualitative evaluations demonstrate that GNet++ significantly outperforms existing baselines by generating human meshes with minimal scene and object penetration, thereby improving the realism and physical accuracy of human-object interactions.

Beyond its immediate use for scene-aware grasping, the MOGRAS dataset serves as a versatile resource for future research in motion generation, simulation, and scene understanding. By providing a rich foundation of physically plausible human-scene interactions, we hope our work encourages the community to explore more complex, dynamic, and context-aware human behaviors in virtual environments.

\bibliography{egbib}

\begin{thebibliography}{42}
\providecommand{\natexlab}[1]{#1}
\providecommand{\url}[1]{\texttt{#1}}
\expandafter\ifx\csname urlstyle\endcsname\relax
  \providecommand{\doi}[1]{doi: #1}\else
  \providecommand{\doi}{doi: \begingroup \urlstyle{rm}\Url}\fi

\bibitem[Bhatnagar et~al.(2022)Bhatnagar, Xie, Petrov, Sminchisescu, Theobalt, and Pons-Moll]{bhatnagar2022behave}
Bharat~Lal Bhatnagar, Xianghui Xie, Ilya~A. Petrov, Cristian Sminchisescu, Christian Theobalt, and Gerard Pons-Moll.
\newblock Behave: Dataset and method for tracking human object interactions, 2022.

\bibitem[Cao et~al.(2020)Cao, Gao, Mangalam, Cai, Vo, and Malik]{cao2020longterm}
Zhe Cao, Hang Gao, Karttikeya Mangalam, Qi{-}Zhi Cai, Minh Vo, and Jitendra Malik.
\newblock Long-term human motion prediction with scene context.
\newblock \emph{CoRR}, abs/2007.03672, 2020.
\newblock URL \url{https://arxiv.org/abs/2007.03672}.

\bibitem[Dai et~al.(2017)Dai, Chang, Savva, Halber, Funkhouser, and Nießner]{dai2017scannet}
Angela Dai, Angel~X. Chang, Manolis Savva, Maciej Halber, Thomas Funkhouser, and Matthias Nießner.
\newblock Scannet: Richly-annotated 3d reconstructions of indoor scenes.
\newblock In \emph{2017 IEEE Conference on Computer Vision and Pattern Recognition (CVPR)}, pages 2432--2443, 2017.
\newblock \doi{10.1109/CVPR.2017.261}.

\bibitem[Delitzas et~al.(2024)Delitzas, Takmaz, Tombari, Sumner, Pollefeys, and Engelmann]{delitzas2024scenefun3d}
Alexandros Delitzas, Ayca Takmaz, Federico Tombari, Robert Sumner, Marc Pollefeys, and Francis Engelmann.
\newblock {SceneFun3D: Fine-Grained Functionality and Affordance Understanding in 3D Scenes}.
\newblock In \emph{IEEE/CVF Conference on Computer Vision and Pattern Recognition (CVPR)}, 2024.

\bibitem[Detry et~al.(2010-01-01)Detry, Kraft, Buch, Krueger, and Piater]{detry2010refining}
Renaud Detry, Dirk Kraft, Anders~Glent Buch, Norbert Krueger, and Justus Piater.
\newblock Refining grasp affordance models by experience, 2010-01-01.
\newblock ISSN 2577-087X.

\bibitem[Dosovitskiy et~al.(2021)Dosovitskiy, Beyer, Kolesnikov, Weissenborn, Zhai, Unterthiner, Dehghani, Minderer, Heigold, Gelly, Uszkoreit, and Houlsby]{vit}
Alexey Dosovitskiy, Lucas Beyer, Alexander Kolesnikov, Dirk Weissenborn, Xiaohua Zhai, Thomas Unterthiner, Mostafa Dehghani, Matthias Minderer, Georg Heigold, Sylvain Gelly, Jakob Uszkoreit, and Neil Houlsby.
\newblock An image is worth 16x16 words: Transformers for image recognition at scale.
\newblock \emph{ICLR}, 2021.

\bibitem[Grady et~al.(2021)Grady, Tang, Twigg, Vo, Brahmbhatt, and Kemp]{patrick2021contactopt}
Patrick Grady, Chengcheng Tang, Christopher~D. Twigg, Minh Vo, Samarth Brahmbhatt, and Charles~C. Kemp.
\newblock Contactopt: Optimizing contact to improve grasps.
\newblock \emph{CoRR}, abs/2104.07267, 2021.
\newblock URL \url{https://arxiv.org/abs/2104.07267}.

\bibitem[Hassan et~al.(2021)Hassan, Ghosh, Tesch, Tzionas, and Black]{Hassan_2021_CVPR}
Mohamed Hassan, Partha Ghosh, Joachim Tesch, Dimitrios Tzionas, and Michael~J. Black.
\newblock Populating 3d scenes by learning human-scene interaction.
\newblock In \emph{Proceedings of the IEEE/CVF Conference on Computer Vision and Pattern Recognition (CVPR)}, pages 14708--14718, June 2021.

\bibitem[Hsiao and Lozano-Perez(2006)]{Hsiao2006ImitationLO}
Kaijen Hsiao and Tomas Lozano-Perez.
\newblock Imitation learning of whole-body grasps.
\newblock \emph{2006 IEEE/RSJ International Conference on Intelligent Robots and Systems}, pages 5657--5662, 2006.
\newblock URL \url{https://api.semanticscholar.org/CorpusID:2468294}.

\bibitem[Jiang et~al.(2021)Jiang, Liu, Wang, and Wang]{hanwen2021generation}
Hanwen Jiang, Shaowei Liu, Jiashun Wang, and Xiaolong Wang.
\newblock Hand-object contact consistency reasoning for human grasps generation.
\newblock \emph{CoRR}, abs/2104.03304, 2021.
\newblock URL \url{https://arxiv.org/abs/2104.03304}.

\bibitem[Jiang et~al.(2024)Jiang, Zhang, Li, Ma, Wang, Chen, Liu, Zhu, and Huang]{jiang2024scaling}
Nan Jiang, Zhiyuan Zhang, Hongjie Li, Xiaoxuan Ma, Zan Wang, Yixin Chen, Tengyu Liu, Yixin Zhu, and Siyuan Huang.
\newblock Scaling up dynamic human-scene interaction modeling.
\newblock In \emph{Proceedings of the IEEE/CVF Conference on Computer Vision and Pattern Recognition}, pages 1737--1747, 2024.

\bibitem[Kalisiak and Panne(2001)]{kalisiak2001animation}
Maciej Kalisiak and Michiel Panne.
\newblock A grasp-based motion planning algorithm for character animation.
\newblock \emph{The Journal of Visualization and Computer Animation}, 12, 07 2001.
\newblock \doi{10.1002/vis.250}.

\bibitem[Karunratanakul et~al.(2020)Karunratanakul, Yang, Zhang, Black, Muandet, and Tang]{korrawe2020graspingfield}
Korrawe Karunratanakul, Jinlong Yang, Yan Zhang, Michael~J. Black, Krikamol Muandet, and Siyu Tang.
\newblock Grasping field: Learning implicit representations for human grasps.
\newblock \emph{CoRR}, abs/2008.04451, 2020.
\newblock URL \url{https://arxiv.org/abs/2008.04451}.

\bibitem[Kim et~al.(2014)Kim, Chaudhuri, Guibas, and Funkhouser]{kim2014shape2pose}
Vladimir~G. Kim, Siddhartha Chaudhuri, Leonidas Guibas, and Thomas Funkhouser.
\newblock Shape2pose: human-centric shape analysis.
\newblock \emph{ACM Trans. Graph.}, 33\penalty0 (4), jul 2014.
\newblock ISSN 0730-0301.
\newblock \doi{10.1145/2601097.2601117}.
\newblock URL \url{https://doi.org/10.1145/2601097.2601117}.

\bibitem[Krug et~al.(2010)Krug, Dimitrov, Charusta, and Iliev]{Krug445289}
Robert Krug, Dimitar Dimitrov, Krzysztof Charusta, and Boyko Iliev.
\newblock On the efficient computation of independent contact regions for force closure grasps.
\newblock In \emph{IEEE/RSJ 2010 International Conference on Intelligent Robots and Systems (IROS 2010) :}, pages 586--591, 2010.
\newblock ISBN 978-1-4244-6675-7.
\newblock \doi{10.1109/IROS.2010.5654380}.

\bibitem[Kry and Pai(2006)]{kry2006synthesis}
Paul~G. Kry and Dinesh~K. Pai.
\newblock Interaction capture and synthesis.
\newblock \emph{ACM Trans. Graph.}, 25\penalty0 (3):\penalty0 872–880, jul 2006.
\newblock ISSN 0730-0301.
\newblock \doi{10.1145/1141911.1141969}.
\newblock URL \url{https://doi.org/10.1145/1141911.1141969}.

\bibitem[Leimer et~al.(2020)Leimer, Winkler, Ohrhallinger, and Musialski]{kurt2020posetoseat}
Kurt Leimer, Andreas Winkler, Stefan Ohrhallinger, and Przemyslaw Musialski.
\newblock Pose to seat: Automated design of body-supporting surfaces.
\newblock \emph{CoRR}, abs/2003.10435, 2020.
\newblock URL \url{https://arxiv.org/abs/2003.10435}.

\bibitem[Li et~al.(2023)Li, Wu, and Liu]{li2023object}
Jiaman Li, Jiajun Wu, and C~Karen Liu.
\newblock Object motion guided human motion synthesis.
\newblock \emph{ACM Trans. Graph.}, 42\penalty0 (6), 2023.

\bibitem[Li et~al.(2007)Li, Fu, and Pollard]{li2007pruning}
Ying Li, Jiaxin~L. Fu, and Nancy~S. Pollard.
\newblock Data-driven grasp synthesis using shape matching and task-based pruning.
\newblock \emph{IEEE Transactions on Visualization and Computer Graphics}, 13\penalty0 (4):\penalty0 732–747, jul 2007.
\newblock ISSN 1077-2626.
\newblock \doi{10.1109/TVCG.2007.1033}.
\newblock URL \url{https://doi.org/10.1109/TVCG.2007.1033}.

\bibitem[Luo et~al.(2024)Luo, Wang, Liu, Zhang, Tessler, Wang, Yuan, Cao, Lin, Wang, Hodgins, and Kitani]{luo2024smplolympicssportsenvironmentsphysically}
Zhengyi Luo, Jiashun Wang, Kangni Liu, Haotian Zhang, Chen Tessler, Jingbo Wang, Ye~Yuan, Jinkun Cao, Zihui Lin, Fengyi Wang, Jessica Hodgins, and Kris Kitani.
\newblock Smplolympics: Sports environments for physically simulated humanoids, 2024.
\newblock URL \url{https://arxiv.org/abs/2407.00187}.

\bibitem[Luo et~al.(2025)Luo, Cao, Christen, Winkler, Kitani, and Xu]{luo2025omnigraspgraspingdiverseobjects}
Zhengyi Luo, Jinkun Cao, Sammy Christen, Alexander Winkler, Kris Kitani, and Weipeng Xu.
\newblock Omnigrasp: Grasping diverse objects with simulated humanoids, 2025.
\newblock URL \url{https://arxiv.org/abs/2407.11385}.

\bibitem[Mahmood et~al.(2019)Mahmood, Ghorbani, Troje, Pons{-}Moll, and Black]{mahmood2019amass}
Naureen Mahmood, Nima Ghorbani, Nikolaus~F. Troje, Gerard Pons{-}Moll, and Michael~J. Black.
\newblock {AMASS:} archive of motion capture as surface shapes.
\newblock \emph{CoRR}, abs/1904.03278, 2019.
\newblock URL \url{http://arxiv.org/abs/1904.03278}.

\bibitem[Pavlakos et~al.(2019{\natexlab{a}})Pavlakos, Choutas, Ghorbani, Bolkart, Osman, Tzionas, and Black]{pavlakos2019smplx}
Georgios Pavlakos, Vasileios Choutas, Nima Ghorbani, Timo Bolkart, Ahmed A.~A. Osman, Dimitrios Tzionas, and Michael~J. Black.
\newblock Expressive body capture: 3d hands, face, and body from a single image.
\newblock \emph{CoRR}, abs/1904.05866, 2019{\natexlab{a}}.
\newblock URL \url{http://arxiv.org/abs/1904.05866}.

\bibitem[Pavlakos et~al.(2019{\natexlab{b}})Pavlakos, Choutas, Ghorbani, Bolkart, Osman, Tzionas, and Black]{vposer}
Georgios Pavlakos, Vasileios Choutas, Nima Ghorbani, Timo Bolkart, Ahmed A.~A. Osman, Dimitrios Tzionas, and Michael~J. Black.
\newblock Expressive body capture: 3d hands, face, and body from a single image.
\newblock In \emph{Proceedings IEEE Conf. on Computer Vision and Pattern Recognition (CVPR)}, 2019{\natexlab{b}}.

\bibitem[Prokudin et~al.(2019)Prokudin, Lassner, and Romero]{sergey2019bps}
Sergey Prokudin, Christoph Lassner, and Javier Romero.
\newblock Efficient learning on point clouds with basis point sets.
\newblock \emph{CoRR}, abs/1908.09186, 2019.
\newblock URL \url{http://arxiv.org/abs/1908.09186}.

\bibitem[Puig et~al.(2018)Puig, Ra, Boben, Li, Wang, Fidler, and Torralba]{Puig2018VirtualHomeSH}
Xavier Puig, Kevin~Kyunghwan Ra, Marko Boben, Jiaman Li, Tingwu Wang, Sanja Fidler, and Antonio Torralba.
\newblock Virtualhome: Simulating household activities via programs.
\newblock \emph{2018 IEEE/CVF Conference on Computer Vision and Pattern Recognition}, pages 8494--8502, 2018.
\newblock URL \url{https://api.semanticscholar.org/CorpusID:49317780}.

\bibitem[Punnakkal et~al.(2021)Punnakkal, Chandrasekaran, Athanasiou, Quiros-Ramirez, and Black]{BABEL:CVPR:2021}
Abhinanda~R. Punnakkal, Arjun Chandrasekaran, Nikos Athanasiou, Alejandra Quiros-Ramirez, and Michael~J. Black.
\newblock {BABEL}: Bodies, action and behavior with english labels.
\newblock In \emph{Proceedings IEEE/CVF Conf.~on Computer Vision and Pattern Recognition (CVPR)}, pages 722--731, June 2021.

\bibitem[Rijpkema and Girard(1991)]{rijpkema1991grasping}
Hans Rijpkema and Michael Girard.
\newblock Computer animation of knowledge-based human grasping.
\newblock In \emph{Proceedings of the 18th Annual Conference on Computer Graphics and Interactive Techniques}, SIGGRAPH '91, page 339–348, New York, NY, USA, 1991. Association for Computing Machinery.
\newblock ISBN 0897914368.
\newblock \doi{10.1145/122718.122754}.
\newblock URL \url{https://doi.org/10.1145/122718.122754}.

\bibitem[Romero et~al.(2022)Romero, Tzionas, and Black]{javier2022embodiedhands}
Javier Romero, Dimitrios Tzionas, and Michael~J. Black.
\newblock Embodied hands: Modeling and capturing hands and bodies together.
\newblock \emph{CoRR}, abs/2201.02610, 2022.
\newblock URL \url{https://arxiv.org/abs/2201.02610}.

\bibitem[Seo et~al.(2012)Seo, Kim, and Kumar]{seo2012wholearm}
Jungwon Seo, Soonkyum Kim, and Vijay Kumar.
\newblock Planar, bimanual, whole-arm grasping.
\newblock In \emph{2012 IEEE International Conference on Robotics and Automation}, pages 3271--3277, 2012.
\newblock \doi{10.1109/ICRA.2012.6225086}.

\bibitem[Shafir et~al.(2023)Shafir, Tevet, Kapon, and Bermano]{priormdm}
Yonatan Shafir, Guy Tevet, Roy Kapon, and Amit~H Bermano.
\newblock Human motion diffusion as a generative prior.
\newblock \emph{arXiv preprint arXiv:2303.01418}, 2023.

\bibitem[Taheri et~al.(2020)Taheri, Ghorbani, Black, and Tzionas]{GRAB:2020}
Omid Taheri, Nima Ghorbani, Michael~J. Black, and Dimitrios Tzionas.
\newblock {GRAB}: A dataset of whole-body human grasping of objects.
\newblock In \emph{European Conference on Computer Vision (ECCV)}, 2020.
\newblock URL \url{https://grab.is.tue.mpg.de}.

\bibitem[Taheri et~al.(2022)Taheri, Choutas, Black, and Tzionas]{taheri2021goal}
Omid Taheri, Vasileios Choutas, Michael~J. Black, and Dimitrios Tzionas.
\newblock {GOAL}: {G}enerating {4D} whole-body motion for hand-object grasping.
\newblock In \emph{Conference on Computer Vision and Pattern Recognition ({CVPR})}, 2022.
\newblock URL \url{https://goal.is.tue.mpg.de}.

\bibitem[Tendulkar et~al.(2023)Tendulkar, Sur\'is, and Vondrick]{tendulkar2022flex}
Purva Tendulkar, D\'idac Sur\'is, and Carl Vondrick.
\newblock Flex: Full-body grasping without full-body grasps.
\newblock In \emph{Conference on Computer Vision and Pattern Recognition ({CVPR})}, 2023.

\bibitem[Wang et~al.(2023)Wang, Lin, Zeng, Luo, Zhang, and Zhang]{wang2023physhoi}
Yinhuai Wang, Jing Lin, Ailing Zeng, Zhengyi Luo, Jian Zhang, and Lei Zhang.
\newblock Physhoi: Physics-based imitation of dynamic human-object interaction.
\newblock \emph{arXiv preprint arXiv:2312.04393}, 2023.

\bibitem[Wang et~al.(2022)Wang, Chen, Liu, Zhu, Liang, and Huang]{wang2022humanise}
Zan Wang, Yixin Chen, Tengyu Liu, Yixin Zhu, Wei Liang, and Siyuan Huang.
\newblock Humanise: Language-conditioned human motion generation in 3d scenes.
\newblock In \emph{Advances in Neural Information Processing Systems (NeurIPS)}, 2022.

\bibitem[Wu et~al.(2022)Wu, Wang, Zhang, Zhang, Hilliges, Yu, and Tang]{wu2022saga}
Yan Wu, Jiahao Wang, Yan Zhang, Siwei Zhang, Otmar Hilliges, Fisher Yu, and Siyu Tang.
\newblock Saga: Stochastic whole-body grasping with contact.
\newblock In \emph{Proceedings of the European Conference on Computer Vision (ECCV)}, 2022.

\bibitem[Xiao et~al.(2024)Xiao, Wang, Wang, Cao, Zhang, Dai, Lin, and Pang]{xiao2024unifiedhumansceneinteractionprompted}
Zeqi Xiao, Tai Wang, Jingbo Wang, Jinkun Cao, Wenwei Zhang, Bo~Dai, Dahua Lin, and Jiangmiao Pang.
\newblock Unified human-scene interaction via prompted chain-of-contacts, 2024.
\newblock URL \url{https://arxiv.org/abs/2309.07918}.

\bibitem[Yang et~al.(2022)Yang, Li, Zhan, Wu, Xu, Liu, and Lu]{yang2022oakinklargescaleknowledgerepository}
Lixin Yang, Kailin Li, Xinyu Zhan, Fei Wu, Anran Xu, Liu Liu, and Cewu Lu.
\newblock Oakink: A large-scale knowledge repository for understanding hand-object interaction, 2022.
\newblock URL \url{https://arxiv.org/abs/2203.15709}.

\bibitem[Zhang et~al.(2020)Zhang, Zhang, Ma, Black, and Tang]{zhang2020place}
Siwei Zhang, Yan Zhang, Qianli Ma, Michael~J. Black, and Siyu Tang.
\newblock Generating person-scene interactions in 3d scenes.
\newblock \emph{CoRR}, abs/2008.05570, 2020.
\newblock URL \url{https://arxiv.org/abs/2008.05570}.

\bibitem[Zhang et~al.(2024)Zhang, He, Wan, Zhang, Deng, Ma, and Wang]{zhang2024diffgraspwholebodygraspingsynthesis}
Yonghao Zhang, Qiang He, Yanguang Wan, Yinda Zhang, Xiaoming Deng, Cuixia Ma, and Hongan Wang.
\newblock Diffgrasp: Whole-body grasping synthesis guided by object motion using a diffusion model, 2024.
\newblock URL \url{https://arxiv.org/abs/2412.20657}.

\bibitem[Zheng et~al.(2016)Zheng, Liu, Dorsey, and Mitra]{zheng2016ergonomics}
Youyi Zheng, Han Liu, Julie Dorsey, and Niloy~J. Mitra.
\newblock Ergonomics-inspired reshaping and exploration of collections of models.
\newblock \emph{IEEE Transactions on Visualization and Computer Graphics}, 22\penalty0 (6):\penalty0 1732--1744, 2016.
\newblock \doi{10.1109/TVCG.2015.2448084}.

\end{thebibliography}

\appendix
\section*{Appendix}

\section{Augmenting Graspable Pelvis Positions in MOGRAS}
\label{sec:augmentation}
For each object and scene receptacle pair in our MOGRAS dataset, we provide not only the generated grasping sequence but also a set of additional pelvis positions from which the object can be feasibly grasped. This augmentation is crucial for training generative models that need to handle a variety of valid starting configurations.

To augment the original pelvis position, we follow a four-step filtering process to identify new, plausible grasping locations:
\begin{enumerate}
    \item \textbf{Initial Sampling:} We randomly sample 5,000 points within a 1-meter radius sphere centered on the object. We discard points that are located above the receptacle or are too far from the ground, as they represent semantically implausible grasping positions.
    \item \textbf{Height Filtering:} We extract the pelvis height from the generated grasping pose and only keep sampled points within a fixed height range of the original pelvis height along the y-axis. This ensures the new positions are at a realistic height for grasping the object.
    \item \textbf{Collision Filtering:} We create a standing-room cuboid centered at each remaining point. By checking for intersection between these cuboids and the scene's geometry, we discard any positions that would result in a human-scene collision.
    \item \textbf{Final Sampling \& Orientation:} From the remaining valid positions, we uniformly sample 10 points to ensure diversity. Each of these points serves as a new, collision-free pelvis position. The orientation of the human at each new position is defined by the vector pointing from the sampled pelvis position to the center of the grasped object, ensuring the human faces the object they intend to grasp.
\end{enumerate}
This process provides a robust set of augmented grasping positions, enriching our dataset for training and evaluation of scene-aware grasping models.

\section{FLEX Modifications}

This section details the technical modifications made to \textbf{FLEX}~\cite{tendulkar2022flex} that were critical for adapting it to generate scene-aware, full-body grasps. Our adjustments were designed to improve both the efficiency and physical plausibility of the generated poses, making the method suitable for large-scale data synthesis.

We introduced the following key changes to the FLEX optimization framework:
\begin{itemize}
    \item \textbf{Reduced Optimization Iterations:} We decreased the number of optimization steps, which significantly improved computational efficiency without compromising grasp quality.
    \item \textbf{Scene-Aware Grasp Pose Selection:} Unlike the original FLEX, which is scene-agnostic, our approach explicitly accounts for environmental constraints during grasp pose selection. This ensures that all generated grasps are physically feasible within the surrounding scene.
    \item \textbf{Post-Optimization Refinements:} We added two refinement steps to enhance the realism and stability of the generated grasping poses:
    \begin{itemize}
        \item \textbf{Wrist Alignment Correction:} This post-processing step ensures a natural wrist orientation by aligning the wrist with the target object, which avoids unrealistic rotations and misaligned contacts.
        \item \textbf{Contact Consistency Check:} We introduced a contact distance threshold to detect and resolve minor misalignments, thereby enforcing stable contact points and preventing compensatory body motions.
    \end{itemize}
\end{itemize}
These modifications allow our approach to produce high-quality, scene-compliant grasps with reduced inference time compared to FLEX's default settings. This makes our method better suited for large-scale dataset generation and real-time applications.

\section{Additional Details of Experiments}
This section provides a comprehensive overview of our experimental setup, detailing the implementation specifics of our method and the adjustments made to baseline methods to ensure a fair comparison.

\subsection{Implementation Details}
\textbf{GNet++ Data Preparation:} GNet++ generates static full-body grasps that avoid scene intersections. For training, we collected all frames from the MOGRAS dataset where the subjects stably grasp objects with their right hand, with no intersection between the human and the scene. This resulted in 11.4k frames for the training set, 1.8k for the testing set, and 932 for the validation set. We did not apply any augmentation during training.

\textbf{GNet++ Architecture:} GNet++ features a cVAE architecture that generates static full-body grasps within a scene, conditioned on the specified object, its location, and the surrounding scene. The encoder first processes the human grasp and encodes it into an embedding space, while a pretrained ViT~\cite{vit}encoder processes the given scene. The decoder then samples from these embeddings and outputs SMPL-X~\cite{pavlakos2019smplx} parameters \((\hat{\Theta})\), head direction \((\hat{\mathrm{q}})\), and hand offset vectors \((\hat{\mathrm{d}}^{\mathrm{h}\rightarrow \mathrm{o}})\). These features are used to refine the predicted SMPL-X parameters, resulting in a more realistic full-body grasp. We trained the model for 30k steps with the Adam optimizer (learning rate: 1e-4, batch size: 128) on a single NVIDIA GeForce RTX 2080 Ti GPU, taking approximately 12 hours.

\textbf{Adjustments to Baseline Methods.} To ensure a fair comparison, we conducted a careful evaluation of the baseline methods \textbf{GOAL}~\cite{taheri2021goal} and \textbf{SAGA}~\cite{wu2022saga}. As these models were originally trained on the \textbf{GRAB}~\cite{GRAB:2020} dataset, they are inherently scene-agnostic. We evaluated them ``out-of-the-box'' on the MOGRAS test set without any fine-tuning to reflect their original design and to directly assess their performance on our scene-aware task. The results from this initial comparison [mentioned in Table 4 and Figure 5 of the main paper] clearly highlight the domain gap and the limitations of these methods in handling scene constraints.

However, to provide a fairer baseline and isolate the impact of our scene-aware components, we also trained a variant of our model, \textbf{GNet++ (w/o scene input)}, on the MOGRAS dataset. This model uses the same architecture as our proposed GNet++ but with the scene-conditioning branch removed. This ablation serves as a strong baseline, as it is trained on the same data as our full model but lacks the ability to reason about the scene. The superior performance of our full GNet++ model over this ablation [Table 4 of the main paper] confirms that our scene-conditioning approach is the primary reason for our improved performance, rather than just a benefit of fine-tuning on MOGRAS.

\subsection{Ground-Truth Contact Annotation}
To define ground-truth contact in the MOGRAS dataset, we use a distance-based threshold. For each frame, we compute the closest distance from every human vertex to the object surface. A human vertex is considered in contact with the object if its distance to the object is less than a predefined threshold of \textbf{2 cm}. This threshold accounts for minor inaccuracies in mesh alignment while ensuring that contact is physically plausible. For the grab data we use the semantic hand annotations as defined in the original GRAB~\cite{GRAB:2020} dataset. Similarly, foot-floor contact is defined as any vertex on the feet or lower leg with a distance to the ground plane \((z=0)\) less than \textbf{2 cm}.

\subsection{Analysis of Contact Precision and Recall}
Our evaluation reports both \textbf{contact precision} and \textbf{recall} to provide a nuanced understanding of model performance.
\begin{itemize}
    \item \textbf{Precision:} A high precision indicates that when a model predicts a contact point, it is likely to be a true positive. Our models, particularly \textbf{GNet++}, achieve high precision, suggesting they are good at avoiding false-positive contacts (e.g., predicting a hand is touching a part of the object it's not actually holding).
    \item \textbf{Recall:} Recall measures the model's ability to find all true contact points. A lower recall indicates that the model is missing some real contact points. This often happens because the model might be too conservative, or it fails to predict contacts that are present in the ground truth but are subtle (e.g., a pinky finger lightly touching the side of an object).
\end{itemize}
In our experiments, the ablation study (Table 4) shows that removing the penetration loss and scene conditioning significantly impacts recall. This suggests that without proper context and collision avoidance, the models struggle to maintain the full, intricate contact points necessary for a stable, realistic grasp.

\subsection{Human Study}
We conducted a human study with 75 participants (38 males, 37 females) to compare the quality of the generated full-body grasps. For each method (GOAL~\cite{taheri2021goal}, SAGA~\cite{wu2022saga}, and GNet++), we rendered 10 human grasp poses in the same scene from the MOGRAS test set. To ensure a fair horizontal comparison, we carefully minimized visual obstructions in the renderings.

Participants were shown 30 images of human grasps synthesized by our method and the baselines. They were asked to rate the full-body grasps on a 1-to-5 scale based on two criteria:
\begin{itemize}
    \item \textbf{Full-body grasp quality:} 1 = poor grasp, 5 = good grasp.
    \item \textbf{Body-scene intersection:} 1 = significant intersection, 5 = no intersection.
\end{itemize}

\begin{table}[ht]
    \centering
    \caption{\textbf{Comparison of Grasp Quality and Scene Intersection Across Methods:} Average participant ratings from a human study comparing the full-body grasp quality (Average Grasp Rating) and the extent of scene intersection (Average Scene Intersection Rating) for GOAL, SAGA, and GNet++ methods on the MOGRAS dataset. Higher scores indicate better performance, with GNet++ achieving the highest ratings in both categories.}
    \label{tab:human}
    \begin{tabular}{l|c|c}
        \toprule
         & \textbf{Average Grasp Rating} & \textbf{Average Scene Intersection Rating} \\
         \midrule
        \textbf{GOAL}~\cite{taheri2021goal} & 3.39 & 3.00 \\
        \textbf{SAGA}~\cite{wu2022saga} & 2.79 & 2.63\\
        \textbf{GNet++} & \textbf{3.42} & \textbf{4.00} \\
        \bottomrule
    \end{tabular}
\end{table}

The average ratings for full-body grasp quality and scene intersection are presented in Table~\ref{tab:human} as the Average Grasp Rating and the Average Scene Intersection Rating, respectively. The results show that users consistently rated GNet++ higher in both criteria, suggesting superior full-body grasp quality and better scene compliance.

\begin{figure*}[ht]
  \centering
  \caption{\textbf{Examples from our proposed dataset showcasing full-body object interaction within 3D indoor scenes.} The samples demonstrate the pre-grasping walking motion, the grasping pose, and the hand-object contact. The dataset captures the intricate interactions between humans and small objects while adhering to the constraints of the surrounding scene, addressing the limitations of existing datasets that either neglect small object interactions or the 3D scene context.}
  \includegraphics[width=\linewidth]{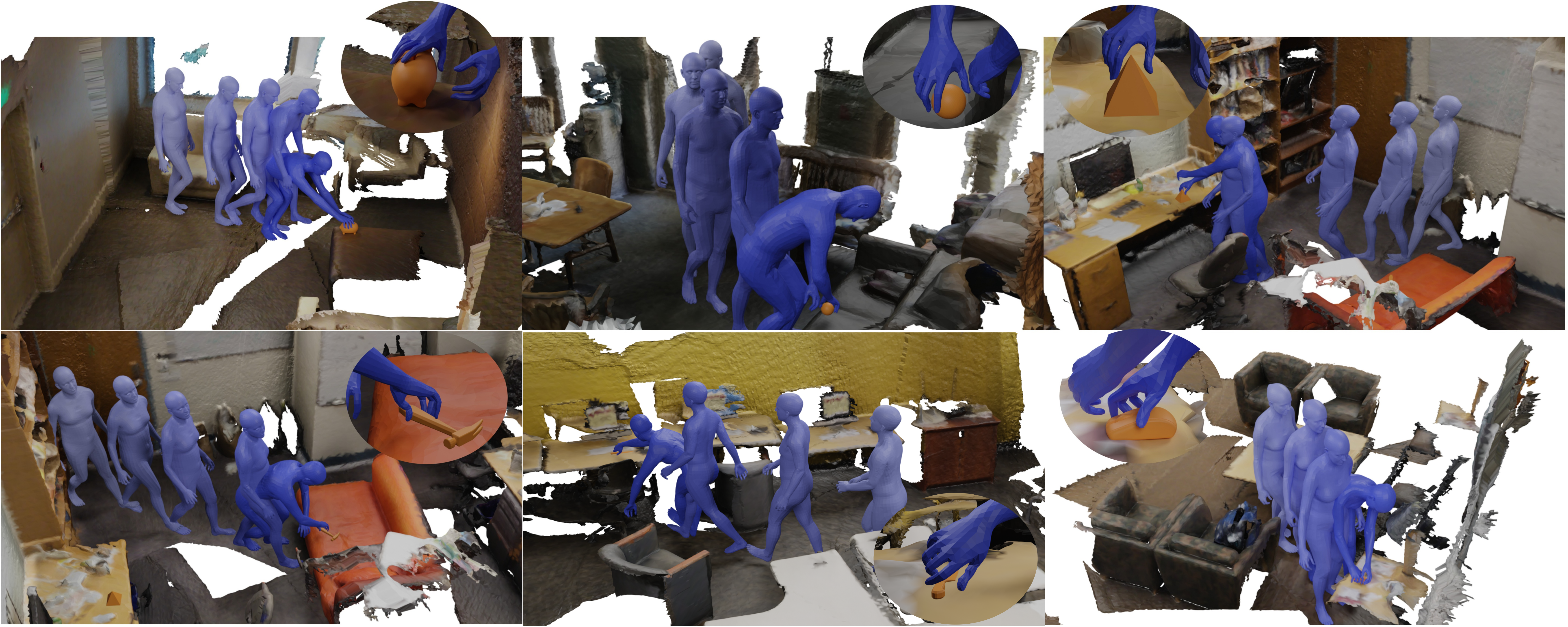}
  \label{fig:data}
\end{figure*}

\end{document}